\documentclass[10pt,journal,compsoc]{IEEEtran}

\ifCLASSOPTIONcompsoc
    \usepackage[nocompress]{cite}
    \usepackage[caption=false,font=footnotesize,labelfont=sf,textfont=sf]{subfig}
\else
    \usepackage{cite}
    \usepackage[caption=false,font=footnotesize]{subfig}
\fi
\ifCLASSINFOpdf
    \usepackage[pdftex]{graphicx}
    \DeclareGraphicsExtensions{.pdf,.jpeg,.png}
\else
    \usepackage[dvips]{graphicx}
    \DeclareGraphicsExtensions{.eps}
\fi
\usepackage{amsmath,amssymb,amsfonts,dsfont,bm,bbm,mathrsfs,pifont}
\usepackage{algorithm,algpseudocode,listings}
\usepackage{booktabs,multirow,adjustbox}
\usepackage{float,capt-of}
\usepackage{url}
\usepackage{xcolor}
\usepackage[pagebackref=false,breaklinks=true,colorlinks=true,citecolor=blue,bookmarks=false]{hyperref}
\usepackage[capitalize]{cleveref}  % Should be loaded after 'hyperref', and works perfectly with 'subfigure'.

\crefname{section}{Sec.}{Secs.}
\Crefname{section}{Section}{Sections}
\crefname{table}{Tab.}{Tabs.}
\Crefname{table}{Table}{Tables}
\crefname{figure}{Fig.}{Figs.}
\Crefname{figure}{Figure}{Figures}
\crefname{equation}{Eq.}{Eqs.}
\Crefname{equation}{Equation}{Equations}
\newcommand{\tabincell}[2]{\begin{tabular}{@{}#1@{}}#2\end{tabular}}
\newcommand{\E}{\mathbb{E}}      % notation of expectation.
\newcommand{\z}{{\rm\bf z}}      % notation of latent code.
\newcommand{\Z}{\mathcal{Z}}     % notation of latent space.
\renewcommand{\t}{{\rm\bf t}}    % notation of transformation code.
\newcommand{\T}{\mathcal{T}}     % notation of transformation space.
\renewcommand{\d}{{\rm\bf d}}    % notation of transformation type.
\newcommand{\x}{{\rm\bf x}}      % notation of image.
\newcommand{\X}{\mathcal{X}}     % notation of image space.
\newcommand{\Loss}{\mathcal{L}}  % notation of loss function.

\begin{document}

%%%% Title
\newcommand{\titlename}{Unsupervised Image Transformation Learning via Generative Adversarial Networks}
\title{\titlename}

%%%% Authors
\author{
  Kaiwen Zha,
  Yujun Shen,
  Bolei Zhou, \IEEEmembership{Member, IEEE}
  \IEEEcompsocitemizethanks{
    \IEEEcompsocthanksitem K. Zha is with the Computer Science and Artificial Intelligence Laboratory, Massachusetts Institute of Technology, Cambridge, MA 02139, USA.\protect\\
    E-mail: kzha@mit.edu
    \IEEEcompsocthanksitem Y. Shen is with the Department of Information Engineering, the Chinese University of Hong Kong, Hong Kong SAR, China.\protect\\
    E-mail: shenyujun0302@gmail.com
    \IEEEcompsocthanksitem B. Zhou is with the Computer Science Department, University of California Los Angeles, Los Angeles, CA 90095, USA.\protect\\
    E-mail: bolei@ucla.edu
  }
  %\thanks{Manuscript received xx/xx/xxxx; revised xx/xx/xxxx.}
}

%%%% Headers
% \markboth{IEEE Transactions on Pattern Analysis and Machine Intelligence}%
% {Shen \MakeLowercase{\emph{et al.}} \titlename}

\IEEEtitleabstractindextext{%

%%%% Abstract
\begin{abstract}
%%%%
In this work, we study the image transformation problem, which targets at learning the underlying transformations (\textit{e.g.}, the transition of seasons) from a collection of unlabeled images.
However, there could be countless of transformations in the real world, making such a task incredibly challenging, especially under the unsupervised setting.
To tackle this obstacle, we propose a novel learning framework built on generative adversarial networks (GANs), where the discriminator and the generator share a \textit{transformation space}.
After the model gets fully optimized, any two points within the shared space are expected to define a valid transformation.
In this way, at the inference stage, we manage to adequately extract the variation factor between a \textit{customizable} image pair by projecting both images onto the transformation space.
The resulting transformation vector can further guide the image synthesis, facilitating image editing with continuous semantic change (\textit{e.g.}, altering summer to winter with fall as the intermediate step).
Noticeably, the learned transformation space supports not only transferring image styles (\textit{e.g.}, changing day to night), but also manipulating image contents (\textit{e.g.}, adding clouds in the sky).
In addition, we make in-depth analysis on the properties of the transformation space to help understand how various transformations are organized.
Project page is at \url{https://genforce.github.io/trgan/}.
\end{abstract}

% Keywords

}

\maketitle
\IEEEdisplaynontitleabstractindextext
\IEEEpeerreviewmaketitle

%%%% Figure: Teaser
\begin{figure*}[t]
    \centering
    \includegraphics[width=0.94\linewidth]{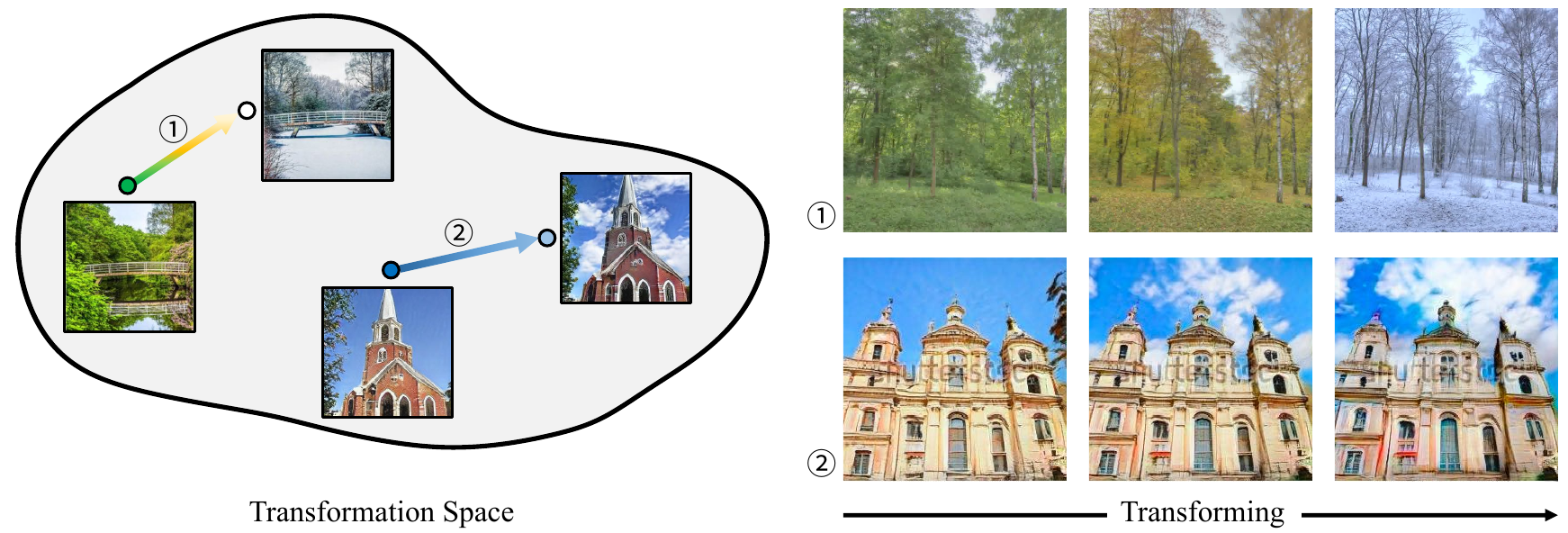}
    \vspace{-10pt}
    \caption{
        Given a customizable image pair, TrGAN is able to extract the semantic variation between them through projecting them onto the learned transformation space.
		Such a variation is applicable to transforming new images.
		Examples on the right show that TrGAN can both transfer image styles (\textit{i.e.}, changing seasons) and, more importantly, edit image contents (\textit{i.e.}, adding clouds).
		In addition, the transformation process is semantically continuous, like yielding fall as the intermediate step when transforming summer to winter.
    }
    \label{fig:teaser}
    \vspace{-5pt}
\end{figure*}

\IEEEraisesectionheading{\section{Introduction}\label{sec:introduction}}

\IEEEPARstart{T}{he} visual world is constantly changing.
The pass of autumn turns the golden color of trees and fields into white, while the sunrise comes to light up the river and sky.
Factors such as season and light drive the visual transformations.
Recent development of deep models has proven to be remarkable in extracting the explanatory factors in a single image~\cite{bengio2013representation}, but the machine still lacks the capability of understanding how images are transformed from one to another.
For example, given a pair of landscape photos, taken in summer and winter respectively, human can not only tell it is the season that varies between the two, but also foresee the effect of season change for a new set of images.
However, it remains challenging for the machine to recognize such an abstract transformation, let alone apply it to transforming new images.

Compared with the conventional classification or generation tasks, it is more difficult to learn transformations.
Firstly, a transformation is often defined by a pair of observations, either of which change may result in a new transformation.
Secondly, there are incredibly large amounts of transformations existing in the visual world, making it impractical to manually label each single one and learn a separate model for it.
Thirdly, transformation usually follows a continuous semantic variation instead of a binary mapping.
A good transformation from one image to another should result in an interpolation effect that is semantically meaningful.
Taking the transition of seasons as an example, changing summer to winter should expect fall as an intermediate step instead of simply turning green color to white.

Towards image transformation learning, existing approaches either involve additional handcrafted labels to define transformation categories~\cite{yang2019semantic,laffont2014transient}, or can only perform limited types of transformations with one model~\cite{isola2017image,zhu2017unpaired,huang2018multimodal,lee2018diverse,choi2020stargan,huang2017arbitrary,li2018closed}.
Consequently, they typically fail to learn the transformations that are not well-defined in the training stage, and hence do not support users in customizing their own transformations with a learned model.

To learn the potential transformations underlying an unlabeled image collection with a unified model, we propose \textit{TrGAN}, which is an unsupervised transformation learning framework based on generative adversarial networks (GANs).
Like other GAN variants, TrGAN employs a generator and a discriminator to compete with each other on the task of image synthesis.
Differently, TrGAN introduces a new \textit{transformation space} beyond the original latent space in GANs, which is shared by the generator and the discriminator.
Concretely, the generator picks a point in the transformation space to produce an individual synthesis, while the discriminator serves as a transformation learner via projecting an image back to a transformation code.
In this way, any two points define a certain transformation.
To reproduce the observed data distribution, the proposed transformation space is expected to model as many transformations as possible, otherwise, the discrepancy between real and fake distributions will be easily spotted by the discriminator.
After the model converges, given an arbitrary image pair for inference, we can use the discriminator to extract the semantic variation between them by embedding both of them to the well-learned transformation space, as shown in \cref{fig:teaser}.
Such variations can be further used to guide the generator for image synthesis, and hence transferred to other samples, enabling versatile image editing.

To summarize, our contributions are as follows.
\begin{itemize}
	\vspace{-2pt}
	\setlength{\itemsep}{2pt}
	\setlength{\parsep}{0pt}
	\setlength{\parskip}{0pt}
	\item We propose TrGAN, by introducing a transformation space shared by the generator and the discriminator, to learn the underlying image transformations from an image collection in an unsupervised manner. During inference, given a customizable pair of images, we are able to adequately extract the variation factor between them.
	\item We are able to apply the semantics that are extracted from any customized image pair to transforming new images. More importantly, the transformation can be interpolated continuously and remain semantically meaningful at the intermediate step, outperforming existing style transfer approaches.
	\item We find that the proposed transformation space is robust to not only support transferring image styles, but also allow editing image contents, as shown in \cref{fig:teaser}. Furthermore, we study the compositionality of various transformation types, which sheds light on the underlying structure of the transformation space.
\end{itemize}

%%%% Section: Related Work
\section{Related Work}
%%%%

\noindent\textbf{Image Transformation.}
Image transformation has been a long-standing topic, whose primary goal is to alter images with certain types of variations.
Some early studies~\cite{reinhard2001color, hertzmann2001image, shih2013data} in this field adopt image analogy to transform images.
Laffont~\textit{et al.}~\cite{laffont2014transient} manually defines 40 transient attributes and employs labeled data for better editing.
Recent advance of neural networks enables high-quality image-to-image translation~\cite{isola2017image,zhu2017unpaired,liu2017unsupervised} and style transfer~\cite{gatys2016image,johnson2016perceptual,huang2017arbitrary,luan2017deep,li2017universal,li2018closed}.
Basically, they borrow the content information from one sample and the style information from another sample (or another domain) to produce a fused image.
This idea is further extended to learn a multi-modal image translator~\cite{zhu2017toward,lee2018diverse,huang2018multimodal,choi2018stargan, meshry2019neural,choi2020stargan}, improving the synthesis diversity.
All these approaches focus on varying the style and appearance of the image while the content remains the same.
Some other work~\cite{patterson2012sun,zhou2014learning} designs scene attributes, which is highly related to the content information, to better characterize the object variations in the images.
Zhan~\textit{et al.}~\cite{zhan2019spatial} particularly studies how to harmoniously add an object to a given image.

TrGAN differs from existing approaches with four main \textbf{improvements}:
\textbf{\textit{(i)}} Unlike prior work~\cite{zhu2017unpaired, zhu2017toward, huang2018multimodal, lee2018diverse, choi2020stargan} that needs to pre-know the transformation type (\textit{e.g.}, certain labels or domains) before training, TrGAN manages to learn underlying transformations without relying on any annotations during training.
For example, if users want to learn the “adding cloud” transformation, CycleGAN~\cite{zhu2017unpaired} and MUNIT~\cite{huang2018multimodal} require to first separate data into “no cloud” and “with cloud” domains before training.
If users want to further learn the “daytime change” transformation, data should be re-separated into “day” and “night” domains and another model is needed.
Similarly, if users want to learn the four seasons within one model, StarGAN2~\cite{choi2020stargan} and DRIT~\cite{lee2018diverse} rely on a dataset that has already been well-labeled with four categories.
By contrast, \textit{TrGAN can achieve all the above goals with a unified framework in an unsupervised learning manner}.
\textbf{\textit{(ii)}} It therefore supports characterizing the transformation from any customized image pair (\textit{i.e.}, users can casually choose their own images of interests) rather than relying on categorical labels or numeric parameters that can only define limited transformation types as in~\cite{yang2019semantic,laffont2014transient}, enabling broader application scenarios.
%and further applying it for image editing.
\textbf{\textit{(iii)}} Instead of transferring style with a simple one-to-one mapping, TrGAN is able to \textit{continuously and semantically} transform images regarding a particular variation.
\textbf{\textit{(iv)}} As shown in \cref{fig:teaser}, TrGAN can not only transfer image styles, but also manipulate image contents, benefiting from the diversity and robustness of the learned transformation space.

\begin{figure*}[t]
	\centering
	\includegraphics[width=0.94\linewidth]{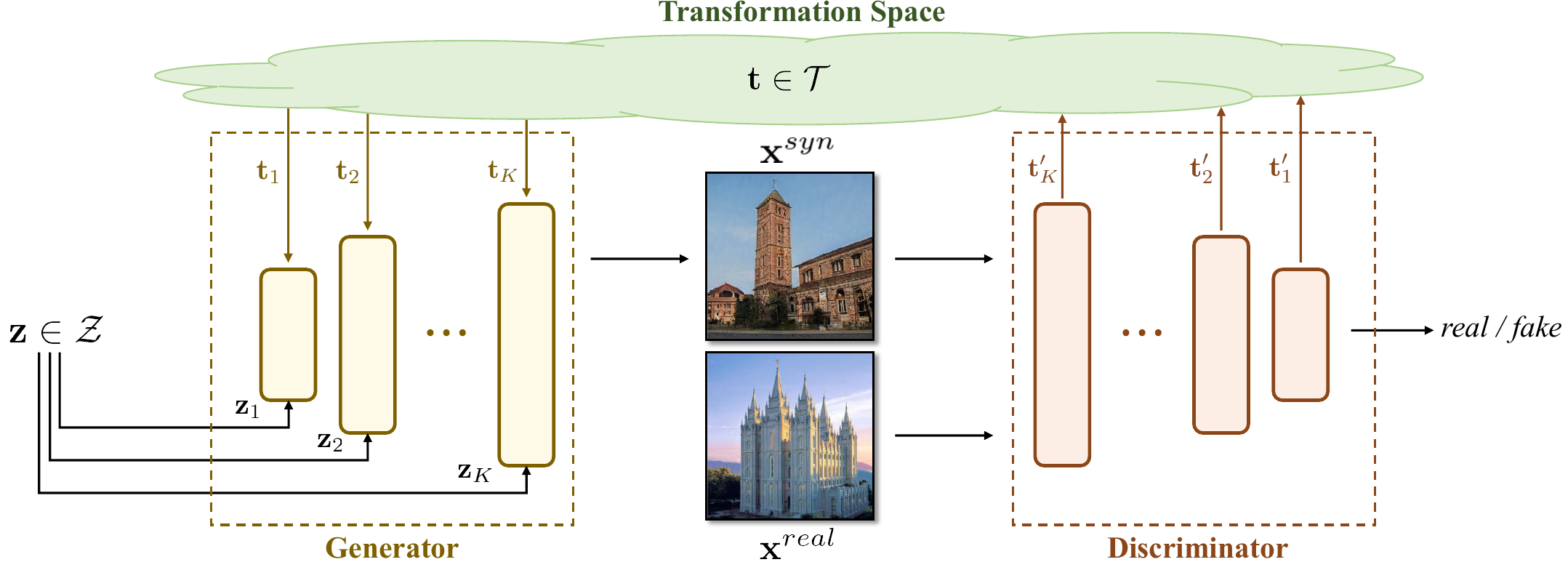}
    \vspace{-10pt}
	\caption{
		\textbf{Concept diagram of TrGAN}.
		In addition to the latent space $\Z$, TrGAN introduces a transformation space $\T$ \textit{shared} by the generator and the discriminator.
		Besides competing with each other on image quality, the discriminator, as the transformation learner, is trained to project any given image to a transformation code, while the generator, as the transformation deployer, learns to employ such a code for controllable synthesis.
		Both the latent code $\z$ and the transformation code $\t$ are organized in a multi-scale manner across layers.
	}
	\label{fig:framework}
	\vspace{-5pt}
\end{figure*}

\noindent\textbf{Generative Adversarial Networks (GANs).}
Recent years have witnessed the advance of GANs~\cite{goodfellow2014generative} in image synthesis~\cite{zhang2019self,brock2019large,karras2017progressive,karras2019style,karras2020analyzing,karras2021alias}.
Starting from a pre-defined latent distribution, GANs model the underlying distribution of the observed data through adversarial training.
It has been recently found that GANs can encode various semantics in the latent space, facilitating image manipulation~\cite{goetschalckx2019ganalyze,jahanian2020steerability,shen2020interpreting,yang2019semantic}.
But they require pre-trained classifier to identify the latent semantics and further use learning-based~\cite{zhu2016generative} or optimization-based~\cite{abdal2019image2stylegan} methods to project an image back to the latent space.
The misalignment between the native latent space and the projected space limits their editing capability to some extent~\cite{zhu2020indomain}.
Instead, TrGAN explicitly introduces a transformation space, which is shared by the generator and the discriminator, to directly learn the underlying transformations from training data.
Meanwhile, the discriminator takes over the inverse mapping from the image space to the transformation space to support extracting variations between customized image pairs.

%%%% Section: Method
\section{TrGAN}
%%%%
\cref{fig:framework} illustrates the framework of TrGAN.
Besides the competition between the generator and the discriminator, we introduce a transformation space $\T$ to bridge them together.
Given a latent code $\z\in\Z$ and a transformation code $\t\in\T$, the generator maps them to a photo-realistic image $\x^{syn}$ that corresponds to a certain point (\textit{i.e.}, $\t$) in the transformation space.
Meanwhile, the discriminator projects $\x^{syn}$ back to the transformation space and gets $\t'$, the approximation of $\t$.
Such design allows the generator and the discriminator to share information.
In this way, we can use the discriminator to extract transformation from any input pair and further apply it to controlling the generator.

\begin{figure*}[t]
	\centering
	\includegraphics[width=0.95\linewidth]{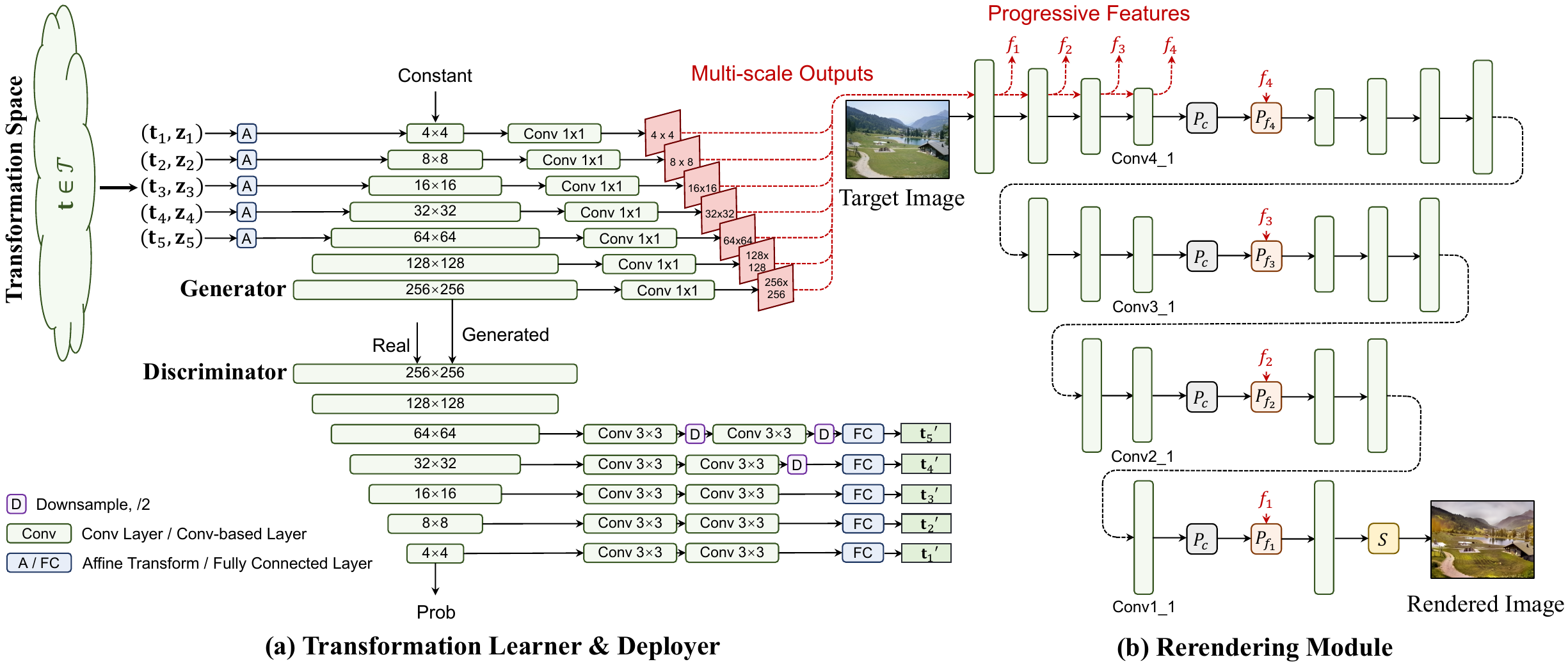}
	\vspace{-10pt}
	\caption{
		\textbf{Detailed architecture of TrGAN.}
		\textbf{Left:} Structures of the transformation learner (discriminator) and the transformation deployer (generator).
		\textbf{Right:} A rerendering module is built upon the transformation deployer to further transform real images.
	}
	\label{fig:supp-framework}
	\vspace{-5pt}
\end{figure*}

\subsection{Shared Transformation Space}
Image transformation can be versatile since one transformation is determined by two images, either of which changing may lead to a completely different transformation.
Let's assume a complete graph with $N$ nodes representing $N$ images respectively.
There are $\frac{N(N-1)}{2}$ edges in total, each of which stands for a particular transformation type.
Along with the dataset growing, the number of transformations will increase dramatically, making it difficult to learn the transformations defined by every single pair.
However, these transformations show clear redundancy.
Taking season changing as an example, any ``summer-winter'' pair should correspond to the same variation.
Hence, to better model the huge diversity and remove the redundancy, we propose to project images onto a transformation space $\T$ instead of directly learning the transformations themselves.
In this way, each image is associated with a particular point in $\T$ and the transformation can be defined by a vector.
More importantly, this transformation space is shared by the generator and discriminator, which unifies the transformation learning and deploying process, and enables us to learn multi-scale transformations.

\subsection{Transformation Learner}\label{sec:learner}
To map the image space to the transformation space, we directly employ the discriminator as the transformation learner considering its encoder structure.
Concretely, we use its intermediate layers to learn the transformation projection, which is shown in \cref{fig:framework}.
Meanwhile, according to the formulation of GANs~\cite{goodfellow2014generative}, the discriminator is also assigned with the task to differentiate the real domain $\X^{real}$ from the synthesized domain $\X^{syn}$.
Like most advanced GAN variants~\cite{brock2019large,karras2019style,karras2020analyzing,shaham2019singan,nguyen2019hologan} that employ layer-wise latent codes, we also train the discriminator to output multi-scale transformation codes to capture more fine-grained characteristics.
Specifically, given an image $\x$, the discriminator with $K$ projection layers will produce
\begin{align}
	D_{adv}(\x) &= p, \\
	D_{T}(\x) &= \t' \triangleq [{\t'_1}^T, {\t'_2}^T, \cdots, {\t'_K}^T]^T,
\end{align}
where $p$ denotes the probability that $\x$ comes from $\X^{real}$ rather than $\X^{syn}$ and $\t'$ is the projected code.

\subsection{Transformation Deployer}\label{sec:deployer}
To better bridge the image space and the transformation space, we expect any point $\t \in\T$ can be projected back to a realistic image, which we call transformation deployment.
This is consistent with the primary goal of the generator.
In other words, on top of taking a randomly sampled latent code $\z\in\Z$ as the input, the synthesis of the generator also depends on a sampled transformation code $\t\in\T$.
Same as the aforementioned transformation projection process, both $\z$ and $\t$ are multi-scale, as $\z \triangleq [\z_1^T, \z_2^T, \cdots, \z_K^T]^T$ and $\t \triangleq [\t_1^T, \t_2^T, \cdots, \t_K^T]^T$.
Then, the images are synthesized through $\x^{syn} = G(\z, \t).$

\subsection{Training Objective}
In addition to the competition between the generator and the discriminator, we would like the transformation learner (discriminator) and the transformation deployer (generator) to share the same transformation space $\T$.
Accordingly, we train the generator and the discriminator with
\begin{align}
	\min_{\Theta_G, \Theta_{D_{T}}} \max_{\Theta_{D_{adv}}}\Loss &=
	\E_{\x \sim \X^{real}}[\log D_{adv}(\x)] \nonumber \\
	&+ \E_{\z \sim \Z, \t \sim \T}[\log(1 - D_{adv}(G(\z, \t)))] \nonumber \\
	&- \lambda\E_{\z \sim \Z, \t \sim \T}[\log P(D_{T}(G(\z, \t))|\t)],
\end{align}
where $\lambda$ is the loss weight.
$P(D_{T}(G(\z, \t))|\t)$ is the approximation distribution of $\t'$ when knowing $\t$.
We maximize the mutual information~\cite{chen2016infogan} between $\t$ and $\t'$ to force the transformation learner and the transformation deployer to share as much information as possible.
Here, we only apply the transformation learner onto synthesized samples since the semantic information contained in real samples is arbitrary under the unsupervised learning setting.

\subsection{Transforming Images}\label{sec:img_trans}
After establishing the mapping from the image space to the transformation space, we are able to extract the semantic variation between any paired data and in turn use the discovered semantics to guide the synthesis process.
More concretely, given an image pair $(\x_A, \x_B)$, we first use the transformation learner to project them onto the transformation space $\T$ with $\t'_A = D_{T}(\x_A)$ and $\t'_B = D_{T}(\x_B)$.
These two points $(\t'_A, \t'_B)$ define a transformation type $\d_{AB} = \t'_B - \t'_A$.
Then we can use the transformation deployer to transform any arbitrary sampled image $G(\z, \t)$, following $T(G(\z, \t)) = G(\z, \t + \gamma\d_{AB})$.
Here, $T(\cdot)$ and $\gamma$ denote the transforming operation and the transforming intensity step respectively.

\subsection{Detailed Framework Structure}

\cref{fig:supp-framework} illustrates the detailed architecture of TrGAN as well as the rerendering module.
Recall that TrGAN introduces a novel transformation space $\T$, which is shared by both the discriminator (transformation learner) and the generator (transformation deployer).
In the generator which starts with a constant feature map~\cite{karras2019style}, the transformation code $\t\in\T$, concatenated with a sampled latent code $\z\in\Z$, is fed into each convolutional layer using Adaptive Instance Normalization (AdaIN)~\cite{huang2017arbitrary}.
Here, we use an affine function $A(\cdot)$ to align the dimension of the combined code $(\t, \z)$ with the number of feature channels in a particular layer.
In the discriminator which aims at differentiating real images from fake ones, two more convolutional layers as well as one more fully-connected layer follow each layer to project the synthesized image back to the transformation space and get $\t'$.
These additional convolutional layers and fully-connected layers refer to $D_{T}(\cdot)$.

\section{Experiments}
In this section, we conduct extensive experiments to demonstrate the effectiveness of TrGAN in learning image transformations.
Before that, we first introduce the datasets used as well as the implementation details.

%%%% Figure: Diverse Transformations
\begin{figure*}
	\centering
	\includegraphics[width=0.98\linewidth]{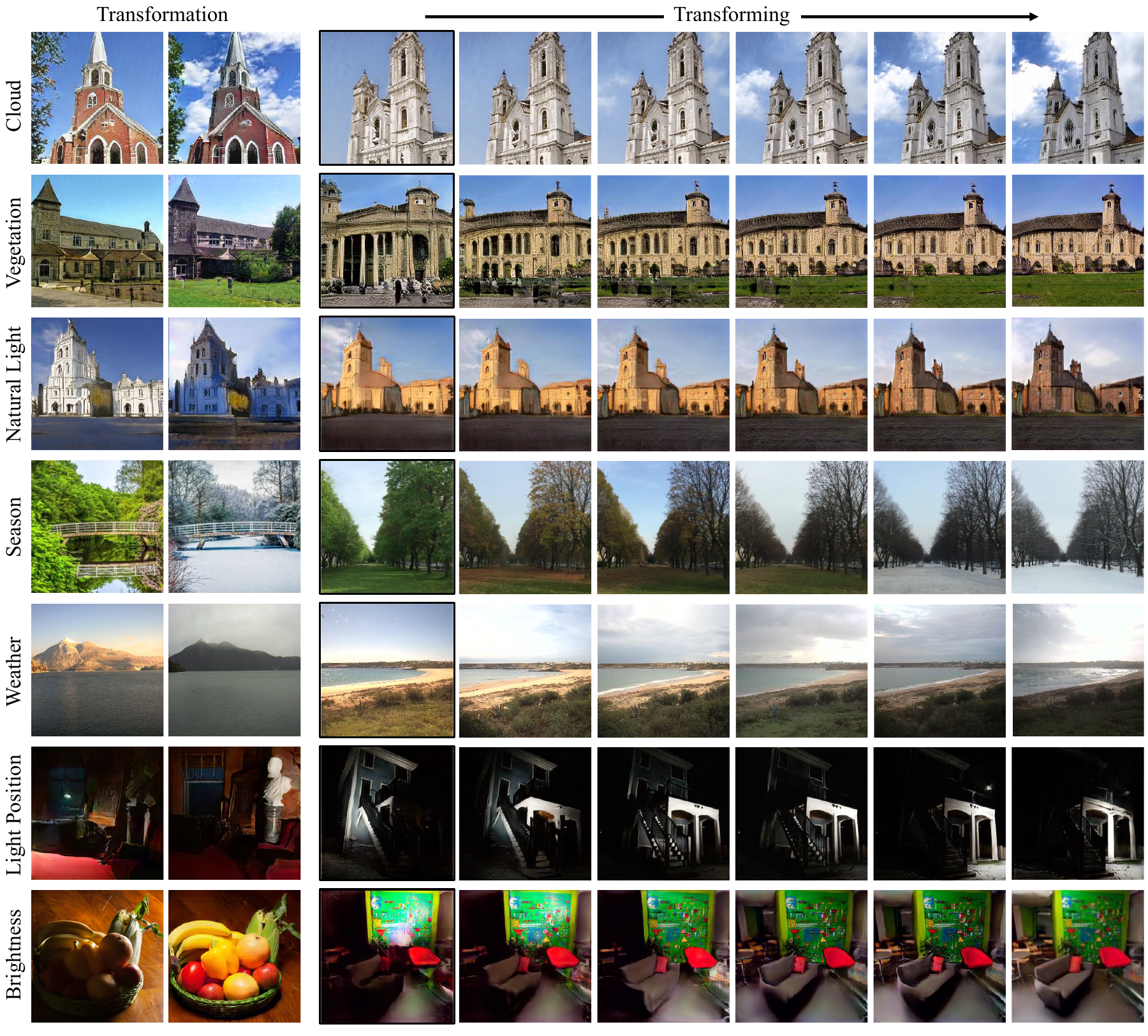}
	\vspace{-15pt}
	\caption{
		\textbf{Diverse transformations} learned by TrGAN.
		The first three rows are from LSUN church, the middle two rows are from our natural scene dataset and the last two rows are from Lighting Compositing dataset.
		The transformations extracted from the image pairs on the left two columns are applied to the images in black boxes.
		The remaining columns show the continuous transforming results.
	}
	\label{fig:all_single_trans}
	\vspace{-5pt}
\end{figure*}

\subsection{Experiment Settings}

\vspace{5pt}
\noindent\textbf{Datasets.}
In our experiments, we validate TrGAN on LSUN church~\cite{yu2015lsun}, Light Compositing dataset~\cite{boyadzhiev2013user}, and a time-lapse natural scene dataset we manually collected from The Webcam Clip Art Dataset~\cite{lalonde2009webcam}, AMOS~\cite{jacobs2009global} and YouTube videos.
For each dataset, we hold out 10\% images as the test set.
\textbf{\textit{(i)}} \textit{LSUN dataset}~\cite{yu2015lsun} is a large-scale scene image dataset consisting of 7 indoor scene categories and 3 outdoor scene categories.
We choose the outdoor church, with $126k$ images in total, as the target set.
\textbf{\textit{(ii)}} \textit{Light Compositing dataset}~\cite{boyadzhiev2013user} contains 6 indoor scene categories: cafe, library, basket, house, sofas, and kitchen, with 112, 83, 129, 149, 32, 127 images respectively.
In each scene (originally dark), the images are captured by a fixed camera with the scene partly lighted up by a moving light source.
Here, we remove those fully lighted-up images.
\textbf{\textit{(iii)}} \textit{Time-lapse natural scene dataset} is manually collected by ourselves from The Webcam Clip Art Dataset~\cite{lalonde2009webcam}, AMOS~\cite{jacobs2009global}, and some YouTube videos.
It contains more than $100k$ natural images with drastically varying appearances.
Here, we shuffle these image sequences into independent images regardless of the temporal correlation.

\vspace{5pt}
\noindent\textbf{Implementation Details.}
We adopt progressive training technique~\cite{karras2017progressive} to train our TrGAN, where the resolution progressively grows from $4\times4$ to $256\times256$.
The transformation codes $\{\t_k\}_{k=1}^{K}$ and the latent codes $\{\z_k\}_{k=1}^{K}$ are injected into the generator from the $4 \times 4$ resolution layer to the $64 \times 64$ layer ($K=5$).
Each transformation code is a 4-D vector, sampled from a uniform distribution $\mathcal{U}(-1, 1)$, while each latent code is a 32-D vector, subject to a normal distribution $\mathcal{N}(0, 1)$.
We use Adam optimizer~\cite{kingma2015adam} to train both the generator and the discriminator.
The learning rate starts from $0.001$ and gradually increases to $0.002$ with the resolution growing.
The loss weight $\lambda$ is set to $1.0$.

%%%% Figure: Comparison to Vector Arithmetic
\definecolor{arylideyellow}{rgb}{0.79, 0.66, 0.22}
\begin{figure*}[t]
	\centering
	\includegraphics[width=0.98\linewidth]{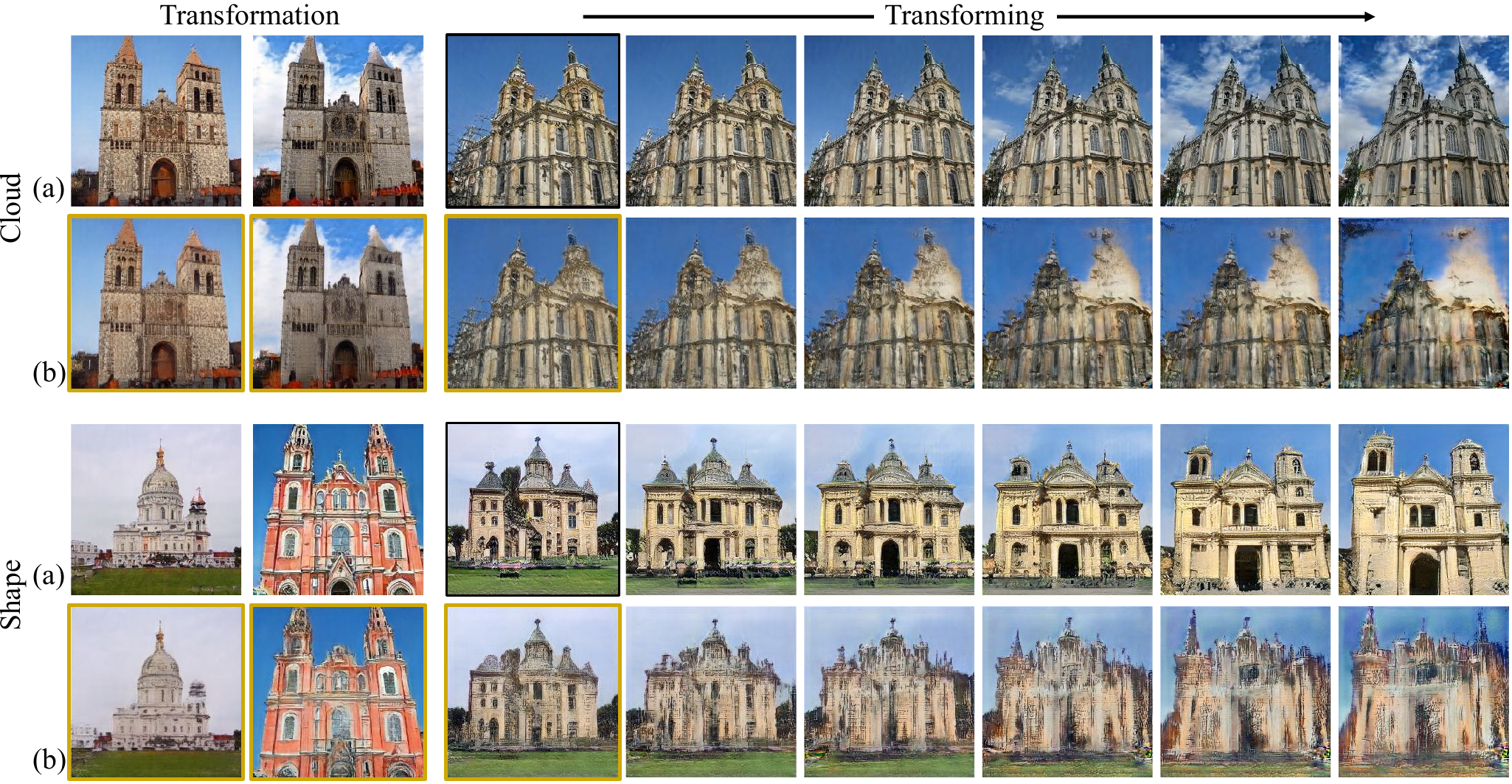}
	\vspace{-10pt}
	\caption{
		\textbf{Qualitative comparison between the transformation space and the latent space.}
		(a) Transforming images by projecting the target image pair (left two columns) to the proposed \textit{transformation space} and using it to manipulate the raw synthesis in black boxes.
		(b) Transforming images by inverting~\cite{abdal2019image2stylegan} the target pair of images to the \textit{latent space} of conventional GANs and performing manipulation through vector arithmetic~\cite{radford2016unsupervised}.
		Samples in yellow boxes indicate the reconstruction results from the inverted codes.
	}
	\label{fig:vector_cmp}
	\vspace{-5pt}
\end{figure*}

\subsection{Learning Transformations}
In this part, we validate the capability of TrGAN in learning underlying transformations from image collections.
We first train models on the three datasets mentioned above and use the transformation learner to project a pair of images, which are not seen in the training process, onto the learned transformation space.
We then adopt the extracted transformation to control the image synthesis, as described in \cref{sec:img_trans}, to see whether TrGAN can properly identify the variation between the inference image pair.

\cref{fig:all_single_trans} shows the results with versatile transformations.
We can tell that TrGAN can handle both the relatively small dataset (\textit{i.e.}, Light Compositing dataset~\cite{boyadzhiev2013user}) and large-scale datasets (\textit{i.e.}, LSUN church dataset~\cite{yu2015lsun} and Time-lapse natural scene dataset) in a completely unsupervised learning manner.
Besides, results reveal that TrGAN is able to adequately extract semantically meaningful transformations from the target pair and further apply them to smoothly transforming a third image accordingly.
For example, TrGAN successfully captures the style-aware variations, such as ``natural light'' (from bright to dim), ``season'' (from summer to fall to winter), ``light position'' (from left to right), and ``brightness'' (gradually lighting up the scene from dark), as shown in \cref{fig:all_single_trans}.
It is also able to identify the content-aware variations, such as ``altering clouds'' and ``adding vegetation'', outperforming existing approaches that are particularly designed for style transfer~\cite{gatys2016image,li2018closed,huang2018multimodal,liu2019few}.
Such a large diversity of transformations are all learned without any annotations, benefiting from the novel transformation space.
In this way, we can discover a great number of transformations from customized paired data with strong robustness.

\noindent\textbf{Ablation on Transformation Space.}
As Radford \textit{et al.}~\cite{radford2016unsupervised} has discovered the vector arithmetic property of the latent space of GANs, a straightforward baseline of TrGAN is using the latent space to replace the transformation space.
In other words, given a target pair of images, we can project them to the native latent space with GAN inversion approaches~\cite{abdal2019image2stylegan,zhu2020indomain} and further adopt vector arithmetic for editing.
\cref{fig:vector_cmp} shows the qualitative comparison where TrGAN demonstrates significant improvement.
For example, in the case of adding clouds, the manipulation in the latent space can only blur the sky part yet fail to alter the image contents reasonably.
Also, in the second example of \cref{fig:vector_cmp}, the latent space does not support extracting the shape variation between the input pair.
By contrast, after separating the transformation space from the latent space, TrGAN can put more effort into the learning of image transformations, with much stronger generalization ability.

\begin{figure*}[!ht]
	\centering
	\includegraphics[width=0.95\linewidth]{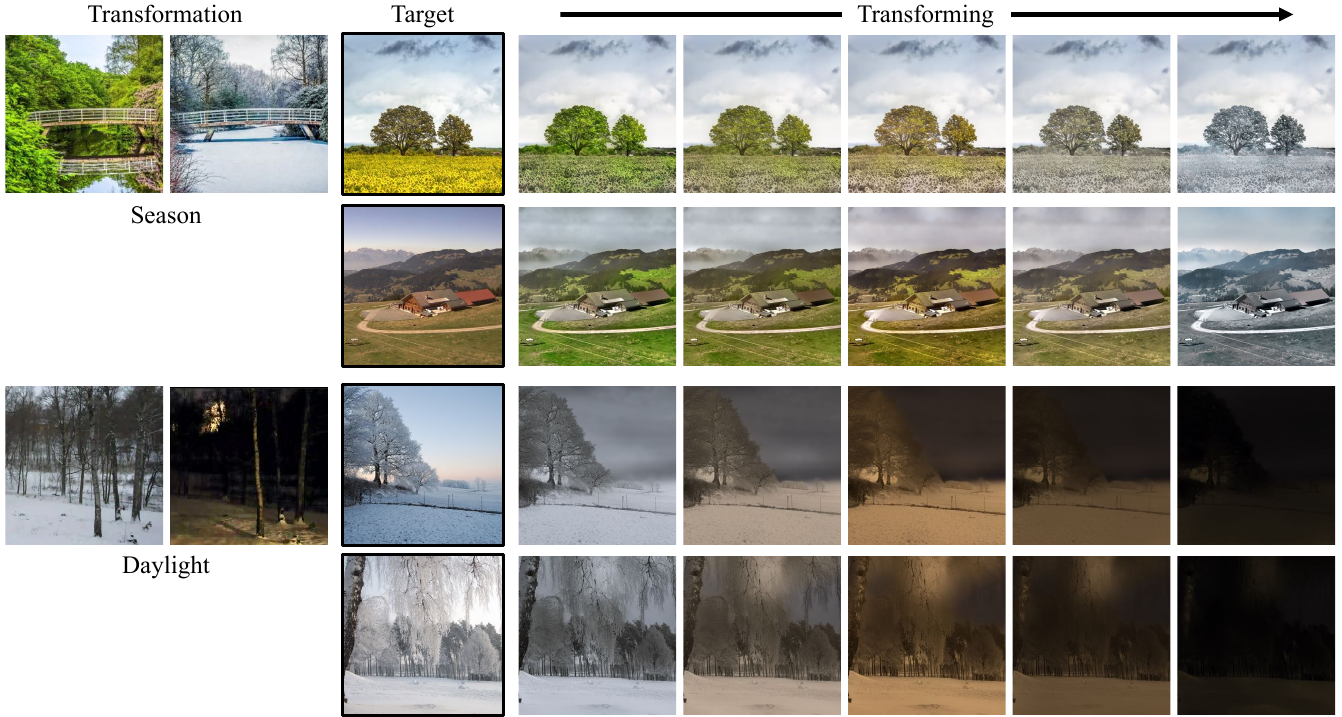}
	\vspace{-10pt}
	\caption{
		\textbf{Transforming season and daylight of real images.}
		For each image group (top two rows and bottom two rows), the input pair is shown on the left, black boxes highlight the target real images, and transforming results are shown on the right.
		Our transformation goes beyond merely interpolating color tones.
		Instead, we can transform season and daylight semantically, \textit{i.e.}, yielding autumn between summer and winter, and producing dusk between morning and night.
	}
	\label{fig:supp-natural_rerender}
	\vspace{-5pt}
\end{figure*}

%%%% Figure: Comparison to style transfer
\begin{figure*}[t]
	\centering
	\includegraphics[width=0.95\linewidth]{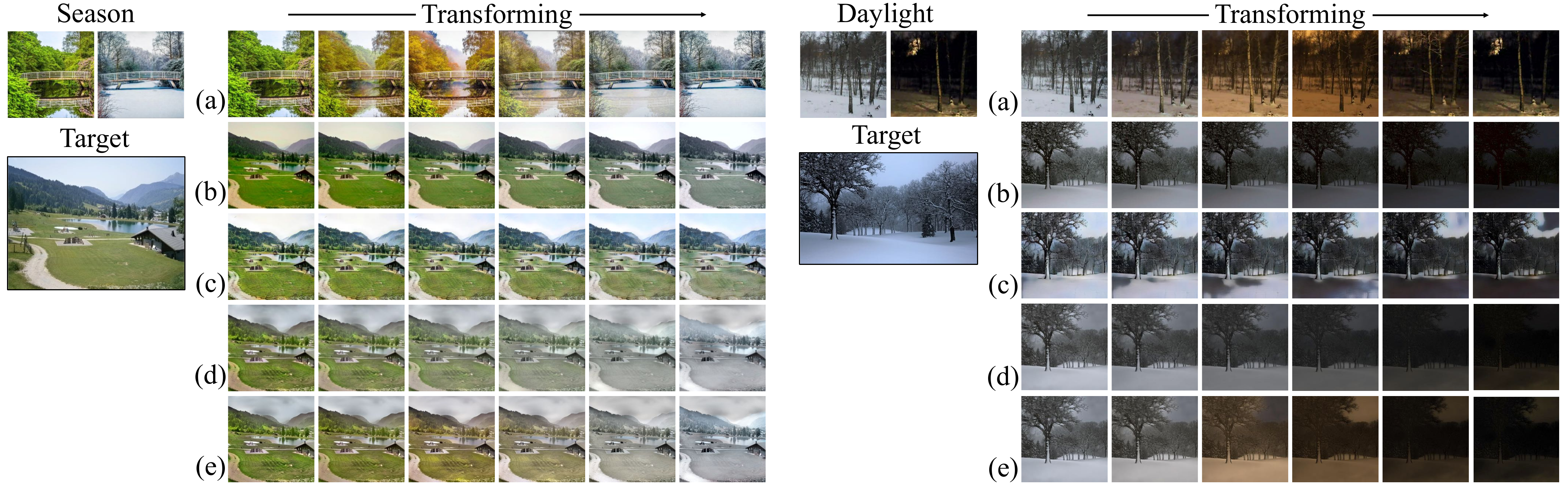}
	\vspace{-10pt}
	\caption{
		\textbf{Qualitative comparison with the style transfer alternatives}.
		For each example, the variation between the top left two images is applied to the target real image.
		Each row shows a sequence of varying images, indicating the gradual transforming process.
		From top to bottom: (a) Ground-truth sequence in the test set; (b) Reinhard~\textit{et al.}~\cite{reinhard2001color}; (c) Huang~\textit{et al.}~\cite{huang2017arbitrary}; (d) Li~\textit{et al.}~\cite{li2018closed}; (e) Ours.
		We beat other competitors, especially at the intermediate steps, by learning a more semantically meaningful transformation.
		Zoom in for details.
	}
	\label{fig:compare_season_daylight}
	\vspace{-5pt}
\end{figure*}

%%%% Figure: Composition
\begin{figure*}[t]
	\centering
	\includegraphics[width=0.95\linewidth]{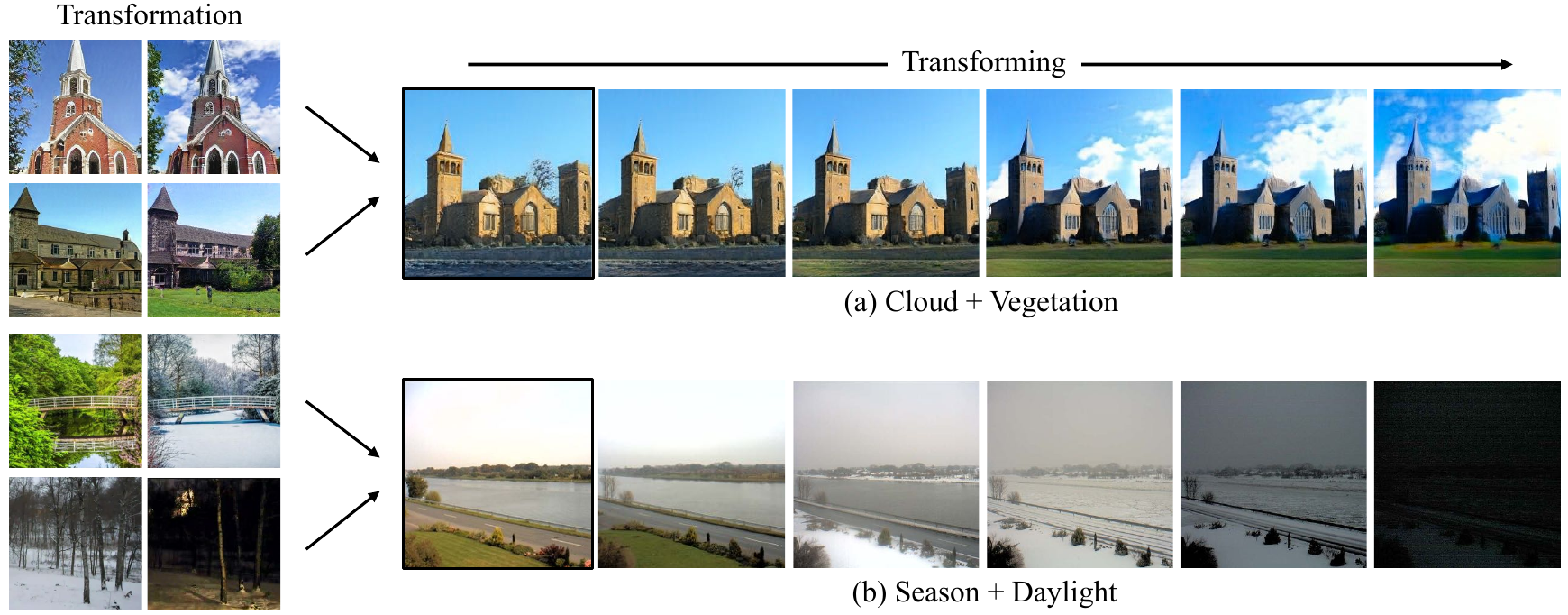}
	\vspace{-10pt}
	\caption{
		\textbf{Composition of two independent transformations}.
		The first two rows on the left are two independent transformations (upper: cloud, lower: vegetation) that modify the image contents, whilst the last two rows on the left are transformations (upper: season, lower: daylight) that change the image styles.
		On the right show the transforming sequences, where the black boxes highlight the base images.
	}
	\label{fig:double_trans}
	\vspace{-5pt}
\end{figure*}

\cref{tab:vector_user} shows the quantitative comparison results.
Here, we pick 10 image pairs as the target transformations.
For each pair, we use the extracted transformation to manipulate 10 images by 5 steps.
Then, we use Frechet Inception Distance (FID)~\cite{fid} as the metric to measure the synthesis quality from each method.
We also conduct a user study on the Amazon Mechanical Turk (AMT) platform to evaluate TrGAN against the baseline approach.
Ten workers are asked two questions for each synthesis, \textit{i.e.}, (i) which method produces images more realistically, and (ii) which method transforms the images more semantically meaningfully.
As suggested in \cref{tab:vector_user}, the novel transformation space is far more competitive than the vanilla latent space.

\subsection{Real Image Editing}\label{sec:real-image}
To enable real image editing with the transformation space, we propose a rerendering module on top of TrGAN.

%%%% Figure: Layer-wise
\definecolor{airforceblue}{rgb}{0.173, 0.439, 0.729}
\definecolor{amber}{rgb}{1.0, 0.93, 0}
\begin{figure*}[t]
	\centering
	\includegraphics[width=0.98\linewidth]{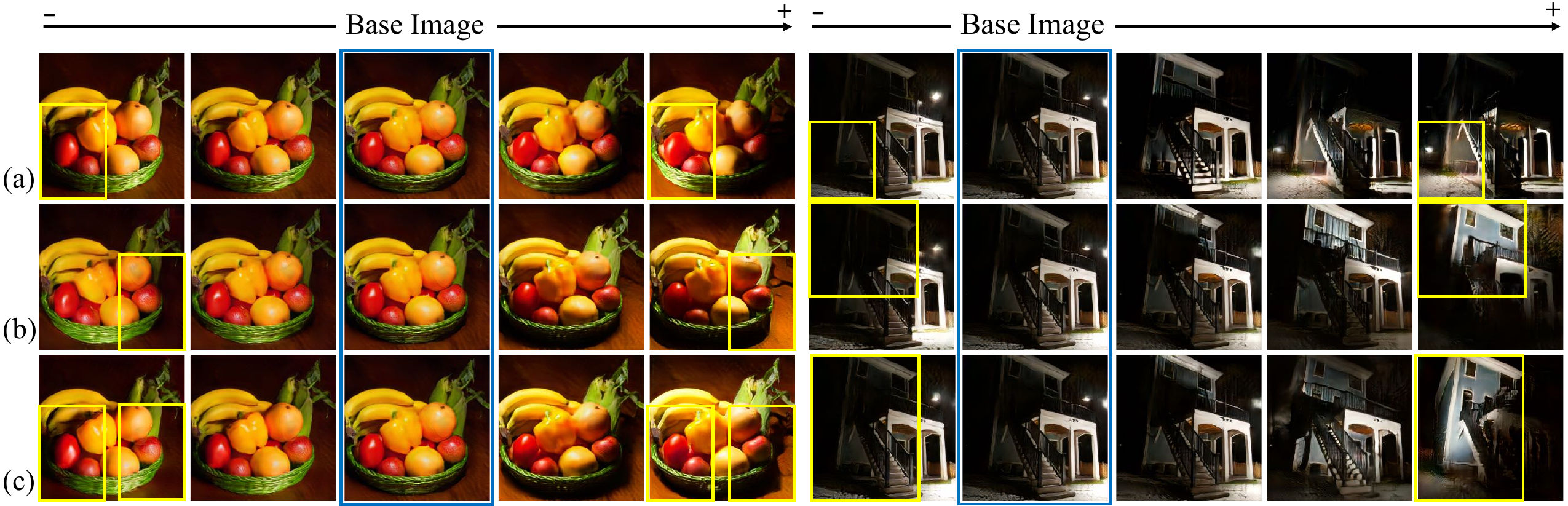}
	\vspace{-10pt}
	\caption{
		\textbf{Layer-wise analysis} of the transformation space.
		%
		% We show the results on two scenes, basket (the left five columns) and house (the right five columns), using the same operations.
		%
		From top to bottom: (a) Transformations by manipulating $\t_1$ (\textit{i.e.}, the transformation code fed into the first layer) along a certain direction upon the base image; (b) Transformations by manipulating $\t_2$ along the same direction; (c) Transformations by jointly manipulating $\t_1$ and $\t_2$.
		Blue boxes highlight the base images.
		Yellow boxes are used to track the changes of lighting conditions.
		We can observe that the transformation codes from different layers tend to control the lighting conditions of different areas.
	}
	\label{fig:layer_wise}
	\vspace{-5pt}
\end{figure*}

%%%% Figure: Unrelated pair
\begin{figure*}[t]
	\centering
	\includegraphics[width=0.98\linewidth]{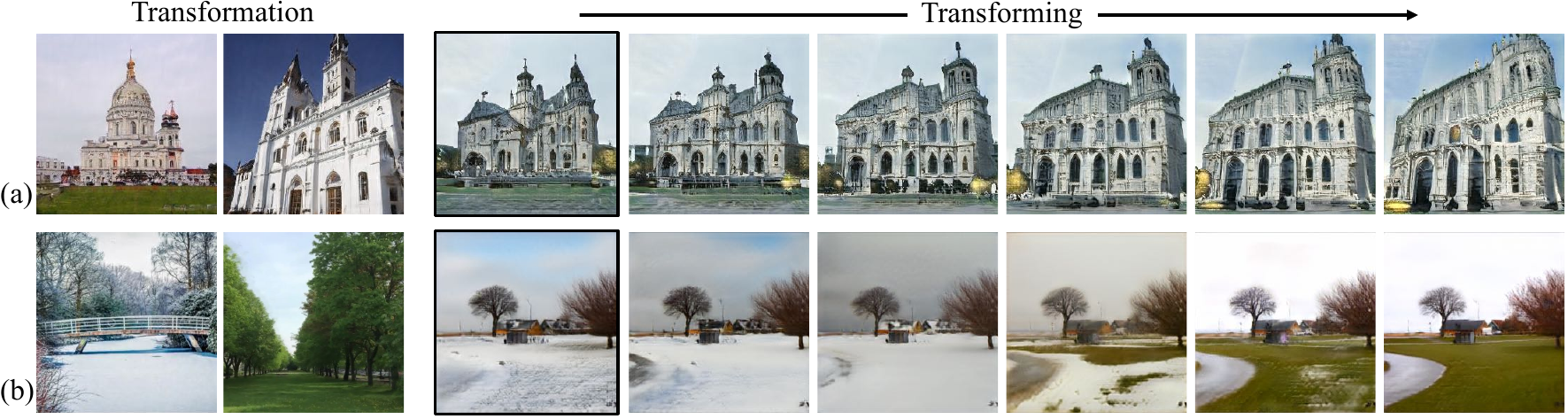}
	\vspace{-10pt}
	\caption{
		\textbf{Transformations discovered from unrelated image pairs}.
		On each row, the left two images are the target image pair that are not related to each other.
		The image sequence on the right shows the results by extracting the transformation from the reference pair and further utilizing it to transform images.
		The black boxes highlight the base images.
		We can see that TrGAN can adequately capture the variations of church shape and season from unrelated pairs, demonstrating the robustness of the learned transformation space.
	}
	\label{fig:unrelated_pair}
	\vspace{-5pt}
\end{figure*}

\vspace{5pt}
\noindent\textbf{Rerendering module.}
Our rerendering module follows an encoder-decoder architecture to edit a real image, which is organized in a multi-level manner, as shown in Fig.~\ref{fig:supp-framework}b.
Different from most style transfer approaches~\cite{gatys2016image,huang2017arbitrary,li2018closed}, which require a style image as input to stylize the target image, our module is designed based on the proposed transformation space.
More concretely, the rerendering module takes the multi-scale intermediate spatial feature maps (\textit{i.e.}, ``multi-scale outputs'' in \cref{fig:supp-framework}) from the transformation deployer, which is explicitly controlled by the layer-wise transformation codes, as the inputs.
These feature maps are all upsampled to the largest scale, and the first-level encoder will extract the ``progressive features'' $\{f_i\}$ from them, as shown in \cref{fig:supp-framework}b.
The encoder and decoder from each level are connected with a whitening function $P_c(\cdot)$ and a coloring function $P_{f_{i}}(\cdot)$~\cite{li2018closed}.
Here, the coloring function $P_{f_{i}}(\cdot)$ is based on an averaged feature map which averages all channels from $f_i$.
Finally, the output from the last-level decoder will be smoothed with a smoothing operation $S(\cdot)$.
The rerendering module is trained from scratch based on the pre-trained generator.

As shown in~\cref{fig:supp-natural_rerender}, our rerendering module can successfully transfer the season and daylight of real images based on the variations extracted from the reference pairs.
More importantly, the transforming process is semantics-aware.
For example, when altering an image from summer to winter, the learned transformation space can properly interpolate a point that corresponds to autumn, instead of simply interpolating color tones by weakening green to white. Similarly, when changing an image from morning to night, the space can produce dusk images, rather than merely interpolating the brightness of the image.

%%%% Table: Comparison to vector arithmetic
\setlength{\tabcolsep}{4.5pt}
\begin{table}[t]
	\caption{
		Quantitative comparison between the novel transformation space proposed in TrGAN and the latent space of conventional GANs.
	}
	\label{tab:vector_user}
	\vspace{-5pt}
	\centering\small
	\begin{tabular}{cccc}
		\toprule[1.2pt]
		& FID
		& \tabincell{c}{More \\ realistic}
		& \tabincell{c}{More semantically \\ meaningful} \\
		\midrule[0.8pt]
		Latent space & 33.69 & 9.4\% & 4.8\% \\
		\textbf{Transformation space} & \textbf{14.57} & \textbf{90.6\%}& \textbf{95.2\%} \\
		\bottomrule[1.2pt]
	\end{tabular}
	\vspace{-10pt}
\end{table}

\vspace{5pt}
\noindent\textbf{Comparison with style transfer alternatives.}
We compare our method with existing style transfer approaches, including Reinhard~\textit{et al.}~\cite{reinhard2001color}, Huang~\textit{et al.}~\cite{huang2017arbitrary} and Li~\textit{et al.}~\cite{li2018closed}.
For style transfer alternatives, we interpolate their style features for continuous transformation.
\cref{fig:compare_season_daylight} and \cref{tab:user} display the qualitative and quantitative comparison results respectively.
We can observe that the transformation obtained by TrGAN is much closer to the ground-truth sequence, especially at the intermediate interpolation steps.
Different from other algorithms that merely interpolate the color tones, TrGAN can produce semantically meaningful results in the continuous transforming process, like fall between summer and winter and dusk between dawn and night, benefiting from the unsupervisedly learned transformation space.
This is also reflected in the user study in \cref{tab:user}.

\subsection{Analysis on the Transformation Space} \label{sec:analysis-of-space}
In this section, we dive deeper into the proposed transformation space.
We first explore its compositionality to see how it performs in combining different transformations.
We then conduct a layer-wise analysis that sheds light on how the transformation space is organized from the layer perspective.
We finally demonstrate the robustness of the transformation space in handling highly unrelated images.

\begin{figure*}[t]
	\centering
	\includegraphics[width=0.95\linewidth]{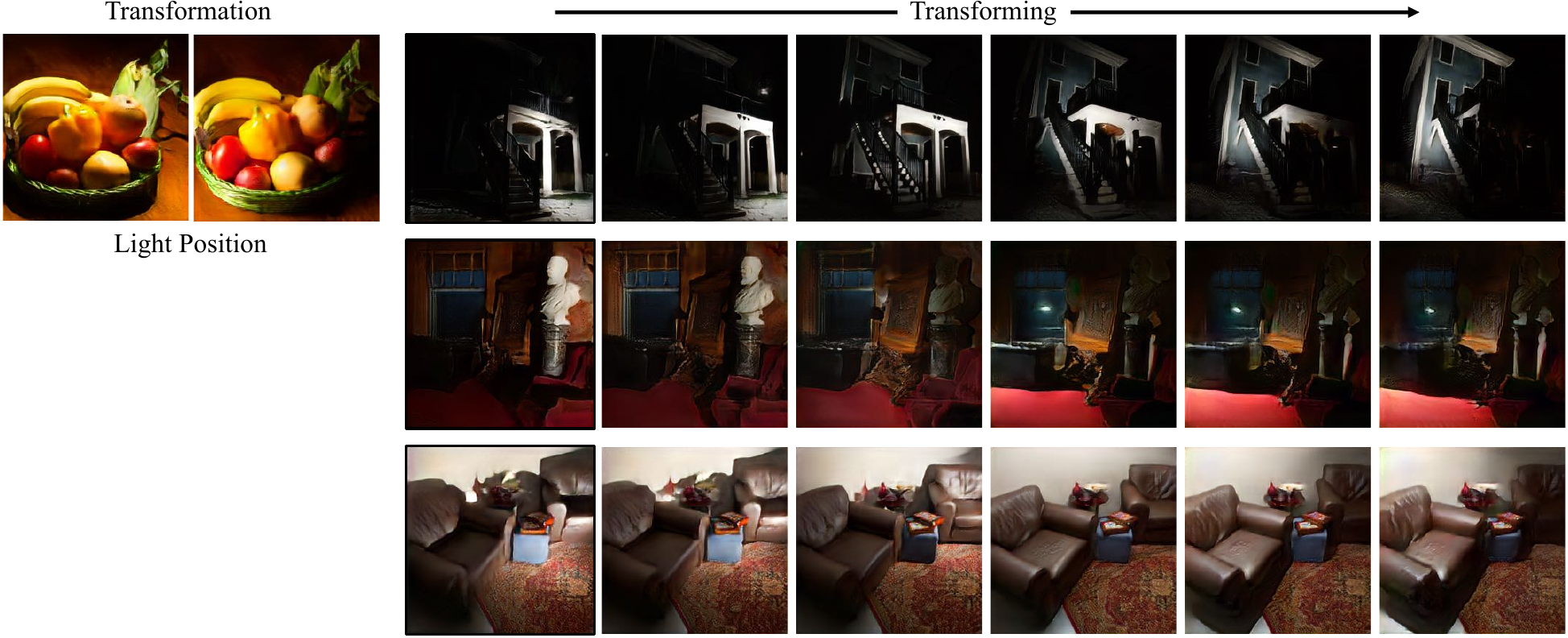}
	\vspace{-10pt}
	\caption{
		\textbf{Altering light position.}
		The input pair is shown on the left (light position from right to left) and transforming samples are on the right.
	}
	\label{fig:supp-light_position}
	\vspace{-5pt}
\end{figure*}

\begin{figure*}[t]
	\centering
	\includegraphics[width=0.95\linewidth]{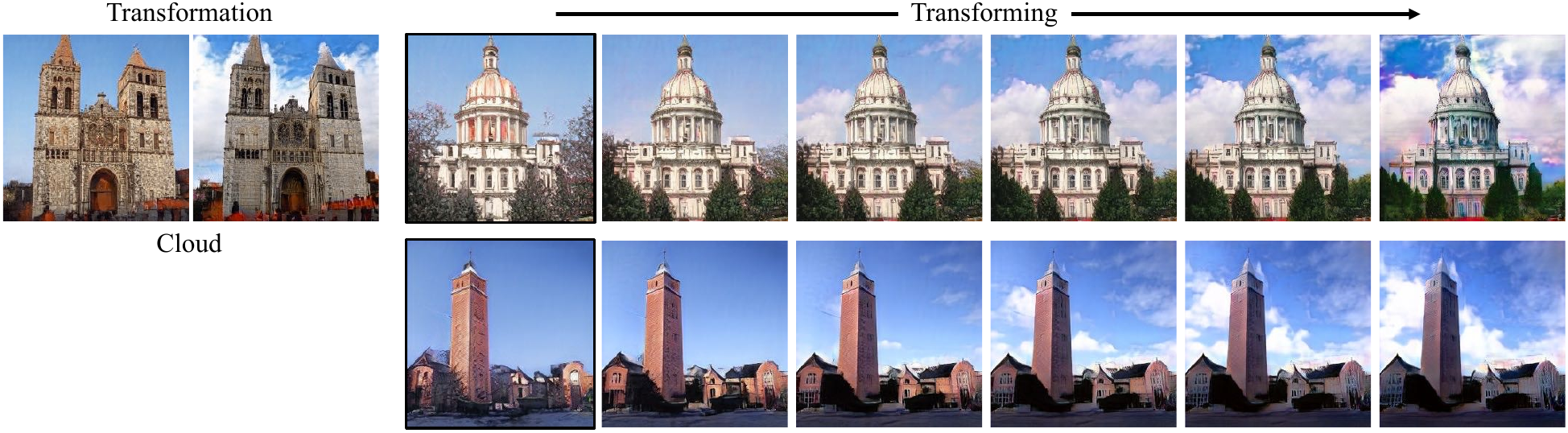}
	\vspace{-10pt}
	\caption{
		\textbf{Adding clouds.}
		The input image pair is shown on the left and transforming samples are on the right.
	}
	\label{fig:supp-cloud}
	\vspace{-5pt}
\end{figure*}

\begin{figure*}[!ht]
	\centering
	\includegraphics[width=0.95\linewidth]{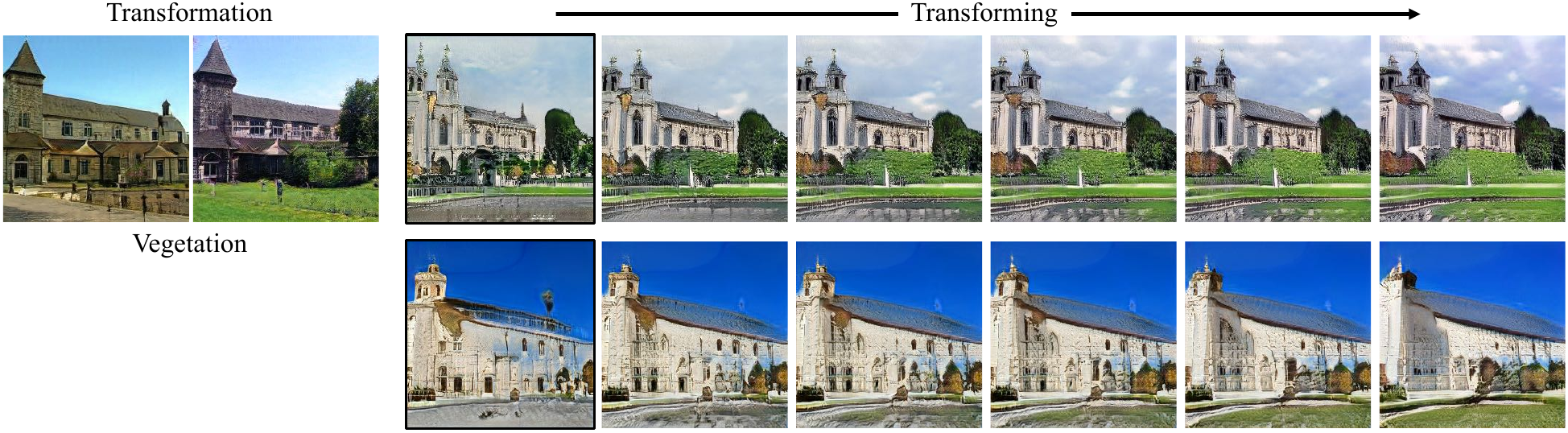}
	\vspace{-10pt}
	\caption{
		\textbf{Adding vegetation.}
		The input image pair is shown on the left and transforming samples are on the right.
	}
	\label{fig:supp-vegetation}
	\vspace{-5pt}
\end{figure*}

\vspace{5pt}
\noindent\textbf{Compositionality of the transformation space.}
Given two pairs of images $(\x_A, \x_B)$ and $(\x_C, \x_D)$, our transformation learner can project them onto the transformation space, getting $(\t'_A, \t'_B)$ and $(\t'_C, \t'_D)$.
Here, we want to validate how the two transformation types $\d_{AB}=\t'_B-\t'_A$ and $\d_{CD}=\t'_D-\t'_C$ can be combined together for a fused image transformation.
\cref{fig:double_trans} gives two examples.
We can tell that transformations learned by TrGAN are flexible for composition.
Taking content editing as an example, we can add clouds and vegetation onto the image simultaneously, as shown in \cref{fig:double_trans}a,
Besides, we can also modulate two style-aware factors (\textit{i.e.}, season and daylight) at the same time, which is presented in \cref{fig:double_trans}b.

\begin{figure*}[t]
	\centering
	\includegraphics[width=0.95\linewidth]{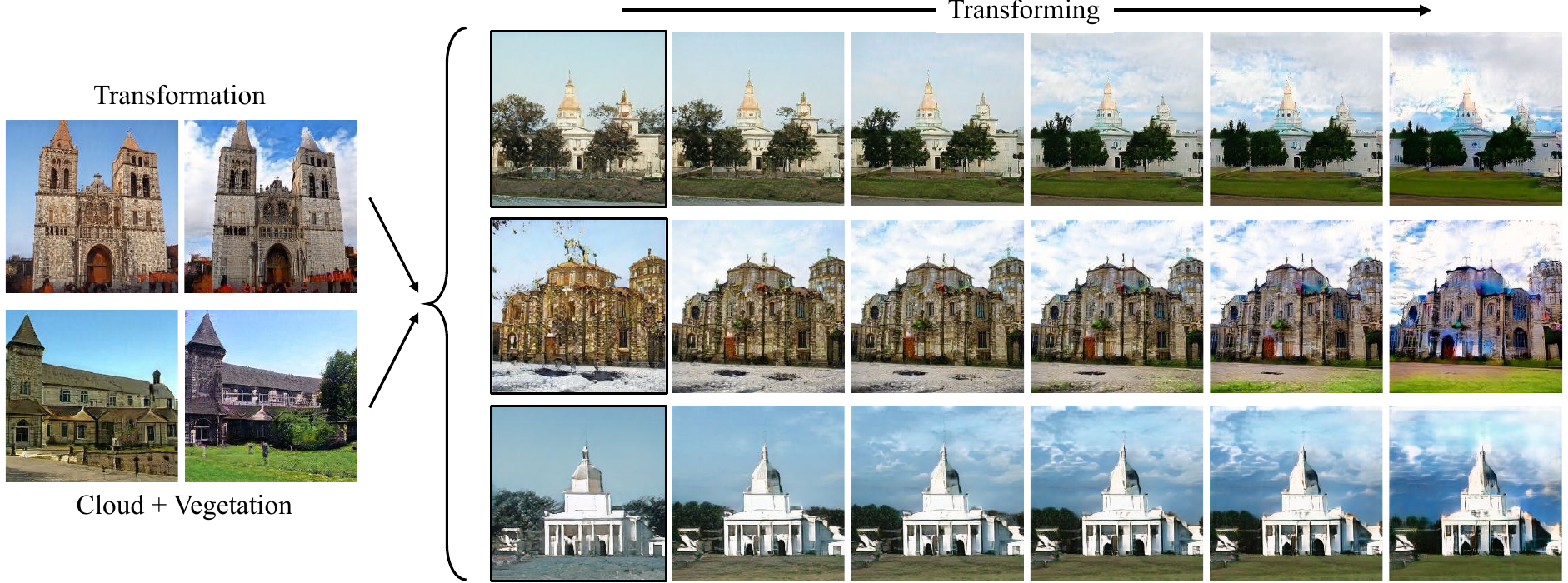}
	\vspace{-10pt}
	\caption{
		\textbf{Simultaneously adding clouds and vegetation.}
		Two independent image pairs (adding clouds, adding vegetation) are embedded to the transformation space and composed together to transform other images.
	}
	\label{fig:supp-cloud_vege}
	\vspace{-5pt}
\end{figure*}

\vspace{5pt}
\noindent\textbf{Disentanglement of the transformation space.}
Recall that both the transformation learner and the transformation deployer employ multi-scale transformation codes.
In this part, we explore how the transformation space is organized across different layers.
In \cref{fig:layer_wise}, we show the results by altering the transformation codes along a certain direction but from different layers.
It turns out that codes at different layers tend to correspond to different transformation controls that are disentangled from each other.
Taking the right sample in \cref{fig:layer_wise} as an example, manipulation at the first layer brightens the bottom left area, while manipulation at the second layer lights up the top left part.
Furthermore, jointly modulating the codes at these two layers can increase the brightness of the entire building on the left.

\begin{figure*}[t]
	\centering
	\includegraphics[width=0.95\linewidth]{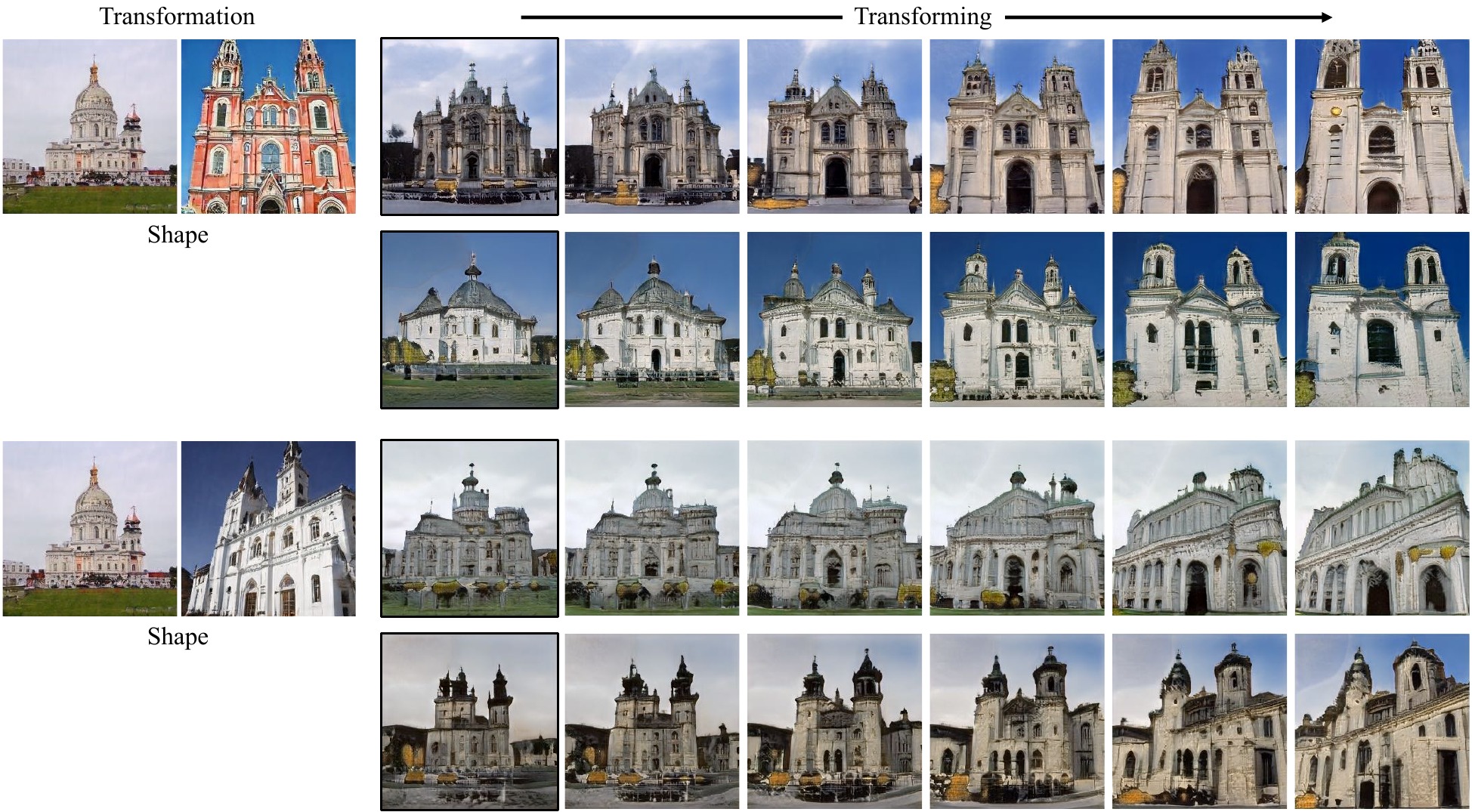}
	\vspace{-10pt}
	\caption{
		\textbf{Transformations discovered from unrelated image pairs.}
		The input pairs on the left are of different church types, shapes and visual angles.
		The transforming samples on the right successfully capture such variations, showing the robustness of the proposed transformation space.
	}
	\label{fig:supp-shape}
	\vspace{-5pt}
\end{figure*}

\vspace{5pt}
\noindent\textbf{Robustness of the transformation space.}
Sometimes, the underlying variation is not easy to spot if the pair images are not vastly related, \textit{e.g.}, a summer picture and a winter picture that are taken at two different places.
Here, we would like to evaluate whether TrGAN can handle such hard cases.
As shown in \cref{fig:unrelated_pair}, our model can still extract semantically meaningful transformations even if the input pair images are highly unrelated.
For example, in \cref{fig:unrelated_pair}a, our model can adequately discover the variations of the church shape as well as the camera angle from the reference pair.
Similarly, in \cref{fig:unrelated_pair}b, the season transformation can be well distilled from the target pair of images regardless of the context difference.
These results verify the robustness of the learned transformation space and show that our method enables broader application scenarios.

%%%% Table: Comparison to style transfer
\setlength{\tabcolsep}{6pt}
\begin{table}[t]
	\caption{
		Quantitative comparison between TrGAN and style transfer alternatives.
	}
	\label{tab:user}
	\vspace{-5pt}
	\centering\small
	\begin{tabular}{cccc}
		\toprule[1.2pt]
		& FID
		& \tabincell{c}{More \\ realistic}
		& \tabincell{c}{More semantically \\ meaningful}\\
		\midrule[0.8pt]
		Reinhard~\textit{et al.}~\cite{reinhard2001color} & 22.78 & 27.0\% & 11.0\% \\
		Huang~\textit{et al.}~\cite{huang2017arbitrary} & 43.66 & 4.0\% & 3.0\% \\
		Li~\textit{et al.}~\cite{li2018closed} & 18.43 & 32.5\%& 14.5\% \\
		\textbf{TrGAN (ours)} & \textbf{17.75} & \textbf{36.5\%}& \textbf{71.5\%} \\
		\bottomrule[1.2pt]
	\end{tabular}
	\vspace{-15pt}
\end{table}

\subsection{More Image Transforming Results}

In this section, we show more image transforming results using TrGAN.
Recall that TrGAN is capable of extracting the transformation from an image pair and further apply such transformation for controllable image synthesis.
It is worth noting that TrGAN can not only transfer image styles (\textit{e.g.}, changing the season of a landscape), but also modulate the objects inside the image (\textit{e.g.}, adding clouds in the sky).
\cref{fig:supp-light_position}, \cref{fig:supp-cloud}, and \cref{fig:supp-vegetation} show the light position, the cloud and the vegetation variations extracted by TrGAN respectively.
The transforming samples suggest that TrGAN can accurately capture the semantic variations between the input pairs and further utilize them to control the synthesis process.
We are even able to compose the transformations extracted from two independent pairs, as shown in \cref{fig:supp-cloud_vege}, validating the compositing property of the learned transformation space $\T$.
Finally, in \cref{fig:supp-shape}, we evaluate TrGAN by taking highly unrelated images as the input pair.
In particular, they are with different church types, as well as different shapes and visual angles.
In this case, the proposed TrGAN can still convincingly discover the meaningful variations between them, verifying the robustness of the learned transformation space and showing that our method enables broader application scenarios.

\section{Limitation, Discussion, and Future work}\label{sec:discussion}

We show that TrGAN can edit image synthesis with learned transformations in Fig.~\ref{fig:all_single_trans} and Fig.~\ref{fig:vector_cmp}. In real-world applications, editing real images is more practical and challenging. To enable it, we have introduced a rerendering module to help transform styles (\textit{e.g.}, changing the season or daylight) of real images in Sec.~\ref{sec:real-image}. But for transformations beyond styles, such as object-related transformations (\textit{e.g.}, adding cloud or vegetation), currently TrGAN is not able to cope with them. One feasible solution in future work is to involve GAN inversion approaches, such as prior work~\cite{abdal2019image2stylegan, zhu2020indomain} that applies well-trained GANs for real image editing, to project the query image onto the original latent space $\Z$ and get $\z'$.
Since we already have the transformation learner to extract the transformation code $\t'$ from it.
We can use $(\t', \z')$ to well reconstruct the target image and further modulate $\t'$ for image transforming.
Another is to re-design the rerendering module and make it applicable to image-to-image translation beyond color tones.
This may require manually labelled paired data for training.
Nevertheless, TrGAN still provides a very promising way to unsupervisedly learn the transformation from image pairs.
It also points out a new direction in transforming images, where users can customize their own transformation pair with no need to re-train the model.

Additionally, during inference, TrGAN tends to extract the most prominent semantic transformation from the query pair if multiple semantic attributes are involved in the transformation, \textit{e.g.}, in Fig.~\ref{fig:vector_cmp} only the most prominent shape transformation instead of other minor transformations like color is captured and used for transforming. However, one huge advantage of our approach is that it supports customizing the transformation pair and also supports taking more than one transformation pairs as the inputs, as shown in Fig.~\ref{fig:double_trans}. Hence, if the user wants to change the color as well, it is possible to take another pair of images, between which only color is different, as the reference. Indeed, being able to capture all potential semantic transformations from a given pair and further disentangle them into different directions is much more ideal, but this requires the transformation learning process to be disentangled. Currently our TrGAN cannot achieve the disentanglement, but we observe that the transformation codes regarding different layers tend to control different kinds of transformations, as specified in Sec.~\ref{sec:analysis-of-space}, which indicates that our well-organized transformation space has already served as a good starting point for the future development of disentangled transformation space.

Another interesting point we've not explored yet is to train TrGAN using a collection of unlabeled image sets from multiple domains. Can TrGAN learn domain-agnostic transformations from this diversified image collection and generalize to transforming images of unseen domains? We leave this as our future work.

%%%% Section: Conclusion
\section{Conclusion}\label{sec:conclusion}
%%%%
In this paper, we present TrGAN to learn the underlying transformations from images in an unsupervised learning manner.
Given an image pair, we can adequately extract the semantic variation between them and further apply it to guiding the synthesis.
Experiment results demonstrate the composition property as well as the versatility of the proposed transformation space in TrGAN.

%%%% Acknowledgment
\ifCLASSOPTIONcompsoc
    \section*{Acknowledgments}
\else
    \section*{Acknowledgment}
\fi
%%%
This work was partly performed when Kaiwen Zha was interning at Computational Perception and Cognition Group, MIT CSAIL. The authors would like to thank Aude Oliva for her support and helpful discussions.

%%%% References
\ifCLASSOPTIONcaptionsoff
  \newpage
\fi
\bibliographystyle{IEEEtran}
\bibliography{ref}

\end{document}